\documentclass[letterpaper, 10 pt, conference]{ieeeconf}  

\IEEEoverridecommandlockouts                              

\usepackage{graphics} 
\usepackage{graphicx} 
\graphicspath{ {Figures/}}
\usepackage{epsfig} 
\usepackage{mathptmx} 
\usepackage{times} 
\usepackage{amsmath} 
\usepackage{amssymb}  
\usepackage{mathtools}
\usepackage{booktabs}
\usepackage{wasysym}
\usepackage{siunitx}
\usepackage{morefloats}
\usepackage[noadjust]{cite}

\usepackage{color,soul}
\usepackage{textcomp}
\usepackage{array}
\usepackage{makecell}
\usepackage{tabularx}
\newcolumntype{K}[1]{>{\centering\arraybackslash}p{#1}}
\usepackage{tabulary}
\usepackage{float}
\usepackage{notoccite}
\usepackage[ruled,vlined]{algorithm2e}
\SetKwInput{KwInput}{Input}
\SetKwInput{KwOutput}{Output}
\usepackage{threeparttable}
\usepackage{nicefrac}
\usepackage[hidelinks]{hyperref}
\usepackage[none]{hyphenat}

\usepackage[table]{xcolor}
\definecolor{blue_background}{HTML}{E3E4FA}
\definecolor{blue_2}{HTML}{E3F2FD}

\title{\LARGE \bf
Design and Preliminary Evaluation of a Torso Stabiliser for Individuals with Spinal Cord Injury
}

\author{Rejin John Varghese$^*$, Man-Yan Tong$^*$, Isabella Szczech, Peter Bryan, Magnus Aronson-Arminoff, \\
Dario Farina, and Etienne Burdet
\thanks{${^*}$equal contribution}
\thanks{This research is supported in part by the UK EPSRC EP/T020970/1 NISNEM and by the EU H2020 REHYB (ICT 871767) grants.} 
\thanks{All authors are with the Department of Bioengineering, Imperial College of Science, Technology and Medicine, London W12 0BZ, UK (email: 
 {\tt\footnotesize \{r.varghese15,e.burdet,d.farina\}@imperial.ac.uk})
 }
}

\begin{document}

\maketitle
\thispagestyle{empty}
\pagestyle{empty}

\begin{abstract}

Spinal cord injuries generally result in sensory and mobility impairments, with torso instability being particularly debilitating. Existing torso stabilisers are often rigid and restrictive. We present an early investigation into a non-restrictive 1 degree-of-freedom (DoF) mechanical torso stabiliser inspired by devices such as centrifugal clutches and seat-belt mechanisms. First, the paper presents a motion-capture (MoCap) and OpenSim-based kinematic analysis of the cable-based system to understand the requisite device characteristics. The evaluation in simulation resulted in the cable-based device to require 55-60\,cm of unrestricted travel, and to lock at a threshold cable velocity of  80-100\,cm/s. Next, the developed 1-DoF device is introduced. The proposed mechanical device is transparent during activities of daily living, and transitions to compliant blocking when incipient fall is detected. Prototype behaviour was then validated using a MoCap-based kinematic analysis to verify non-restrictive movement, reliable transition to blocking, and compliance of the blocking.

\indent \textit{Clinical relevance}— This work establishes the need for torso stabilisation in individuals with Spinal Cord Injury (SCI) and proposes a low-cost passive platform to achieve the same.
\end{abstract}


\section{Introduction} 
Spinal cord injury (SCI) has been increasingly recognised as a global health priority, with statistics varying between $12.1$ and $57.8$ cases per million in high-income countries, and between $12.7$ and $29.7$ in low-income countries \cite{barbiellini2022epidemiology}.
Injuries, especially to the cervical and thoracic regions, compromise torso stability that can severely affect an individual's ability to perform activities of daily living (ADL) such as reaching tasks that require extending one's torso either forwards or sidewards. Additionally, torso stability also affects the ability to perform bimanual tasks as one hand needs to be constantly used to support the torso \cite{britten2018effect}. Furthermore, an inability to maintain a balanced sitting posture also increases the risk of developing pressure sores, which can lead to complications such as autonomic dysreflexia \cite{pressure_sore}. An assistive device designed to provide on-demand support across various seated postures would fill an important open medical need and substantially improve the quality of life of patients.

Several studies have highlighted the importance of external support for individuals with spinal cord injuries to maintain proper sitting posture and prevent abnormal spinal curvature \cite{CD_1,CD_2}. Common solutions include passive devices such as lateral trunk supports (LTS), thoracolumbosacral orthoses (TLSO), and specially equipped wheelchairs. Lateral trunk support is often used clinically to improve seated posture in individuals with SCI who have difficulty in maintaining proper alignment \cite{CD_3}. However, its rigid design can be restrictive, can cause discomfort if not positioned correctly, and may shift out of place. TLSO support the mid-to-lower back but can limit bending and twisting, and are not always suitable for SCI patients, due to risks like pressure sores. Wheelchair modifications like seat-tilting offer stability but lack support for forward/lateral movement and activities.

\begin{figure}
    \centering
    \includegraphics[width=0.85\linewidth,trim=8 4 8 12,clip]{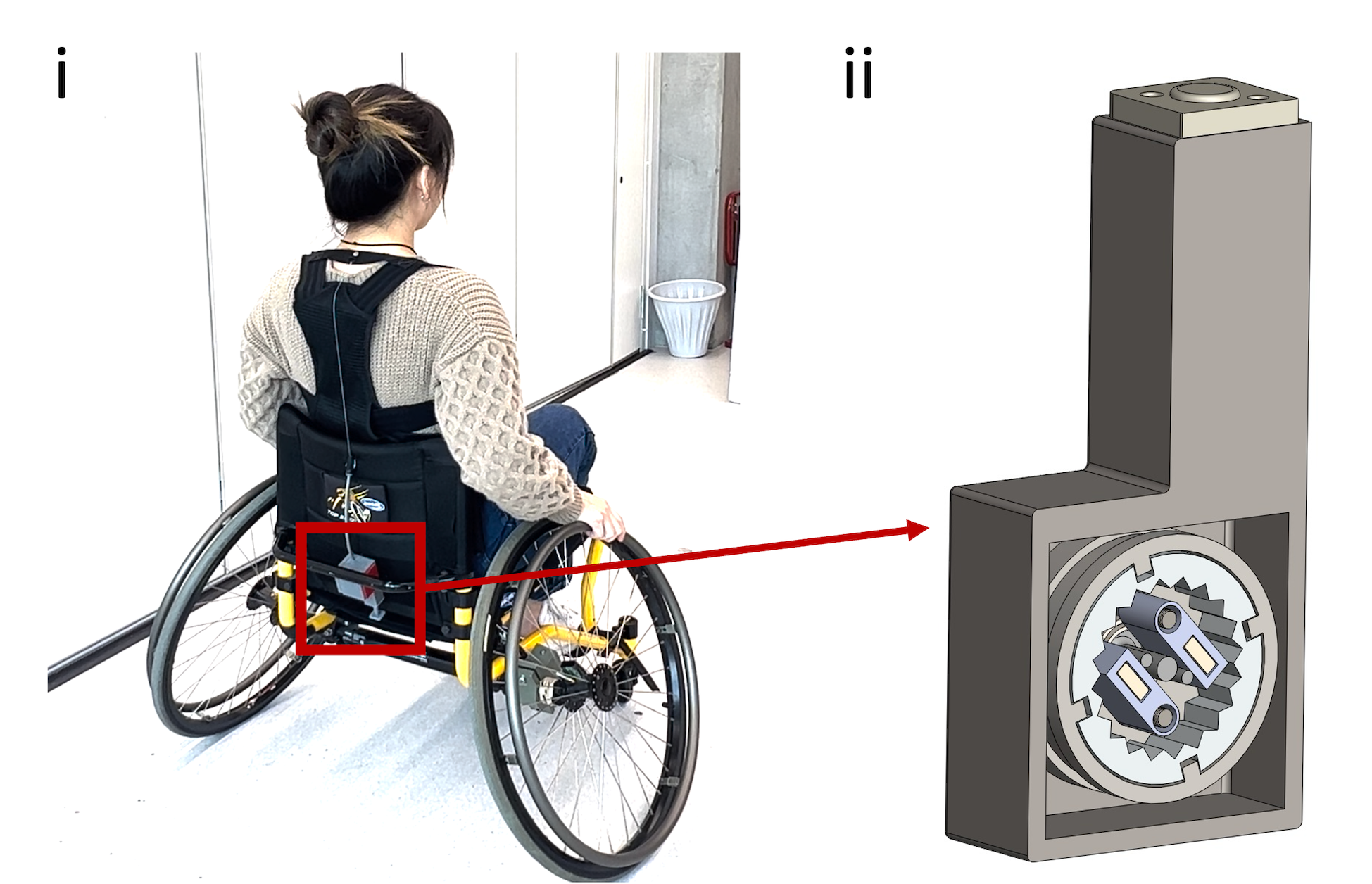}
    \caption{The proposed torso stabiliser system. (i) incorporated as a device attached to the wheelchair, and the cable from the device anchored to a back-support vest. (ii) Section view of the developed prototype.}
    \label{fig:summaryPic}
\end{figure}

To address these limitations in state-of-the-art devices, we investigated a novel torso stabiliser concept (Fig.\,\ref{fig:summaryPic}). 
The proposed device provides a purely mechanical solution designed to offer transparency during ADL and responsive but compliant on-demand blocking upon detecting a fall. The choice for pursuing a mechanical solution was to ensure robustness, simplicity, cost-effectiveness, and, thereby, possible adoption in low- and middle-income countries. The developed prototype is capable of blocking only a single degree-of-freedom (DoF) but can in the future be extended to two DoFs. In this paper, we identify device design parameters, present an early version of the concept, and perform a preliminary validation of the system.

\begin{figure*}
    \centering
    \includegraphics[width=0.9\linewidth,trim=4 4 4 8,clip]{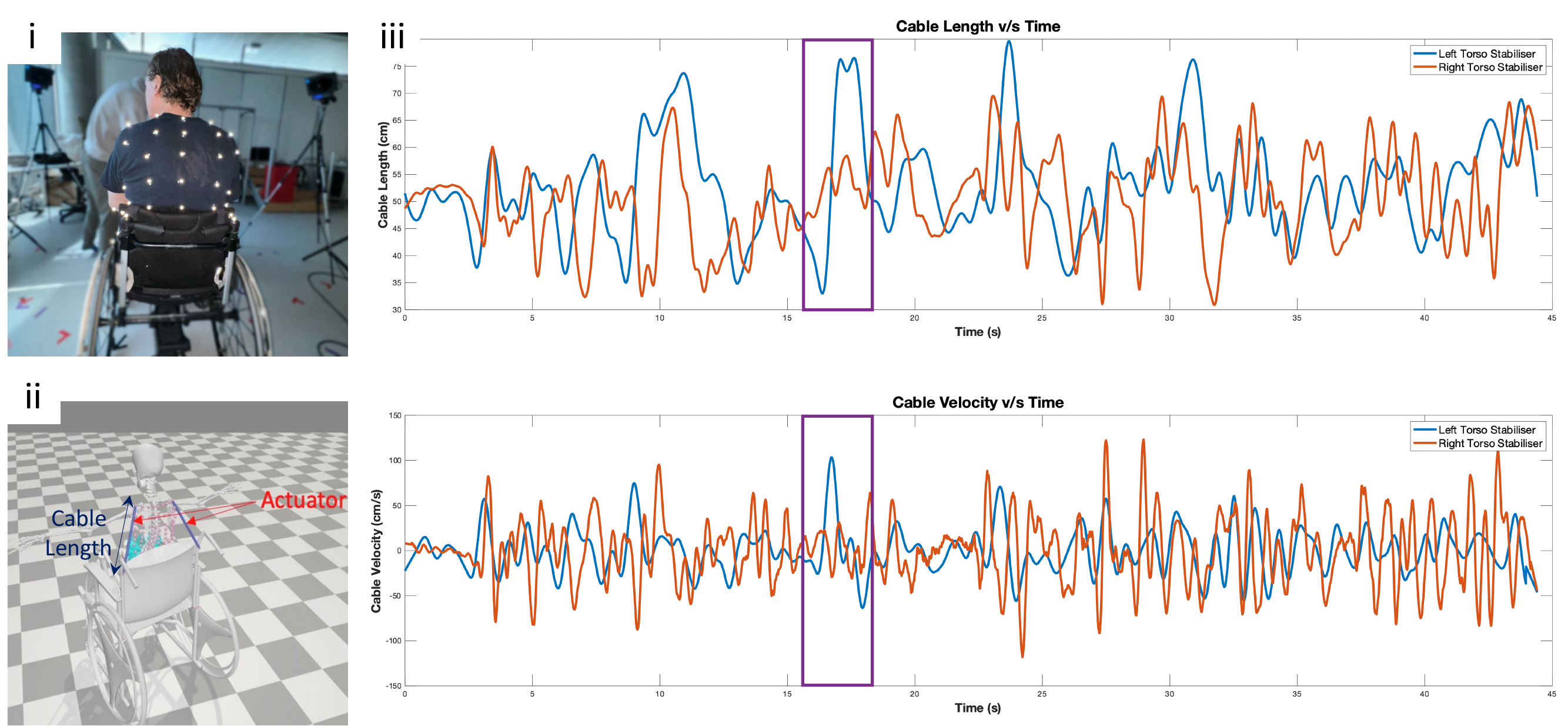}
    \caption{Overview of the virtual prototyping setup and results: (i). Setup for MoCap experiments, (ii). Musculoskeletal model in OpenSim with path actuators, (iii). Torso stabiliser cable displacement and velocity (filtered) v/s time during a trial (blue line - left device, and orange line - right device).}
    \label{fig:problemQuantification}
\end{figure*}

\section{Simulation-based Parameter Identification}

To develop a cable-based device capable of being transparent/non-restrictive during ADL, but promptly blocking a fall, it is critical to identify parameters such as unrestricted cable travel, and a kinematic signature that can be used to trigger the blocking mechanism. For this purpose, MoCap experiments of typical ADL scenarios were simulated with a patient having a T1-T4 level SCI. The data was used to generate a subject-specific musculoskeletal model in OpenSim \cite{delp2007opensim}. Signed informed consent was obtained from the patient before the study.

\subsection{MoCap Experiments}
A multi-camera MoCap system (Smart DX, BTS Bioengineering, Italy) was used for capturing movement. A total of 65 markers were placed both on anatomical features as well as over the back in a grid pattern (Fig.\,\ref{fig:problemQuantification}i). This was done to capture both kinematic movements and the shape of the back for optimising device anchor placement in the future. The entire spine movement could not be tracked as the wheelchair backrest obscured markers on the lower spine from the cameras. Data was captured at $250\,\si{\hertz}$.

Adequate safety protocols were put in place before commencing experiments. A total of eight trials were conducted. Two trials were static data capture trials that would be used for scaling the musculoskeletal model. The remaining six trials captured forward and sideward leaning experienced while simulating ADL reaching activities. These trials were used to quantify range of motion and identify a kinematic signature observed on incipient fall.

\subsection{OpenSim-based Musculoskeletal Modelling}
For our analysis, we used an open-source personalisable full-body model with a detailed thoracolumbar spine (\href{https://simtk.org/projects/pers_fbm_spine}{link}) developed by Schmid \textit{et al.} \cite{schmid2022skin}. Marker locations on the model corresponded to those of the MoCap experiments. Markers on anatomical features from the static trials were used to scale the model to the subject, with RMSE obtained being well inside the acceptable limits in the field. Two path actuators were added to the MS model to mimic the proposed device's behaviour, with a very small tension force applied to ensure the actuators remained taut and followed the subject's movements. The resulting model is shown in Fig.\,\ref{fig:problemQuantification}ii. Simulations using the MoCap data from the kinematic trials were performed using the OpenSim-MATLAB API.
 \begin{figure*}
    \centering
    \includegraphics[width=0.85\linewidth,trim=4 4 4 4,clip]{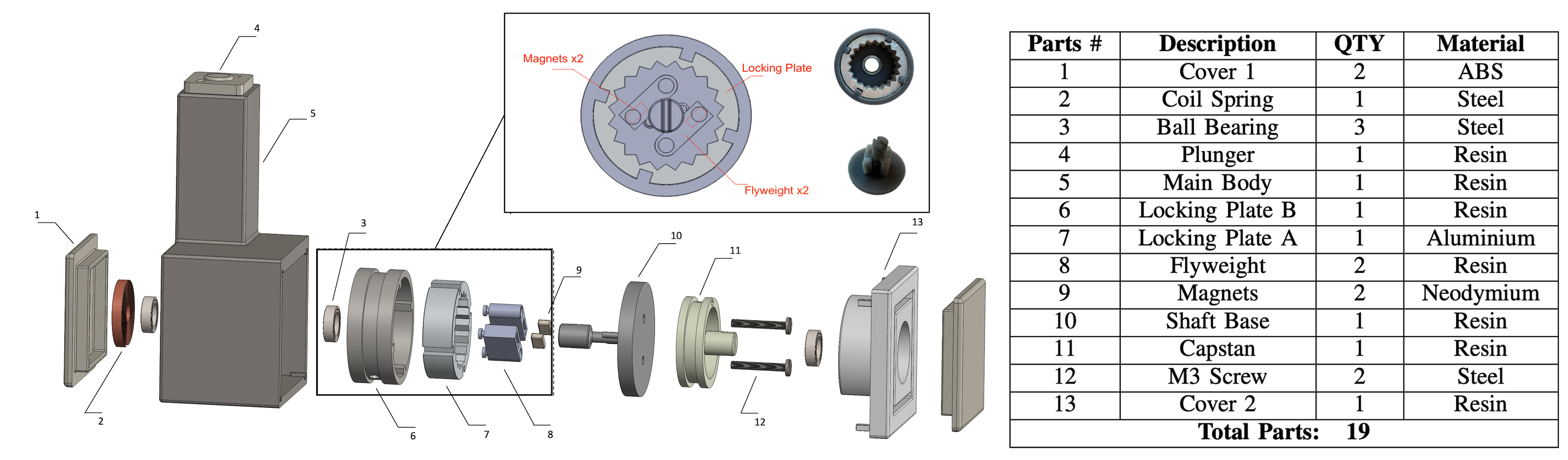}
    \caption{An exploded view and list of components of the developed device that enables both non-restrictive ADL movement and responsive compliant blocking. The mechanism that triggers compliant blocking is presented in the inset figure.} 
    \label{fig:deviceConcept}
\end{figure*}

\subsection{Simulated Evaluation Results}
\label{subsec:simEval}
MoCap data of the trials' movement were used as input to the model developed in OpenSim. 
While performing inverse kinematics simulations in OpenSim, as markers could not be placed all down the spine, the kinematics of the spine could not be accurately reconstructed. However, the movement of the shoulder, where the device was anchored  (Fig.\,\ref{fig:problemQuantification}ii), was accurately tracked.
Fig.\,\ref{fig:problemQuantification}iii shows the results from a trial with both sideward and forward reaching movements, to simulate picking up objects from the ground. The orange and blue lines depict the cable length and cable velocity of each simulated device (path actuator). Cable length data was filtered using the Savitzky-Golay filter in the Signal Analyser app in MATLAB to obtain smoothed cable velocity data. Parts of the trial that depict both the actuators moving in unison represent forward and backward movements, whereas individual actuators moving out of sync represent movements with a sideward component. The simulated evaluation resulted in the device to require $55-60\,\si{\centi\meter}$ of unrestricted travel.

Between $16-18\,\si{\second}$, the subject experienced an incipient fall and reflexively stabilised himself by holding onto the wheelchair frame. In this time window, we observed a particular pattern in the kinematics, consistent with observations reported in literature. Luvstrek \textit{et al.} \cite{luvstrek2011detecting} and Cao \textit{et al.} \cite{cao2016fall} have both reported that at the onset of the fall, acceleration data (measured using IMUs) have a significant drop below the baseline of 1g $(9.8\,\si[per-mode=symbol]{\meter\per\second^2})$, followed by a spike which culminates in the fall. Extrapolating the same to velocity and displacement data (and removing the effect of gravity), we note the kinematic pattern of a rapid forward movement followed by a sharp backward movement, consistent with a reflex and compensation to prevent falling. Based on the above, we identified a cable linear velocity threshold of $80-100\,\si[per-mode=symbol]{\centi\meter\per\second}$ to be translated into equivalent capstan angular velocity as required to trigger the blocking mechanism.
\begin{figure*}
    \centering
    \includegraphics[width=0.9\linewidth,trim=8 8 12 4,clip]{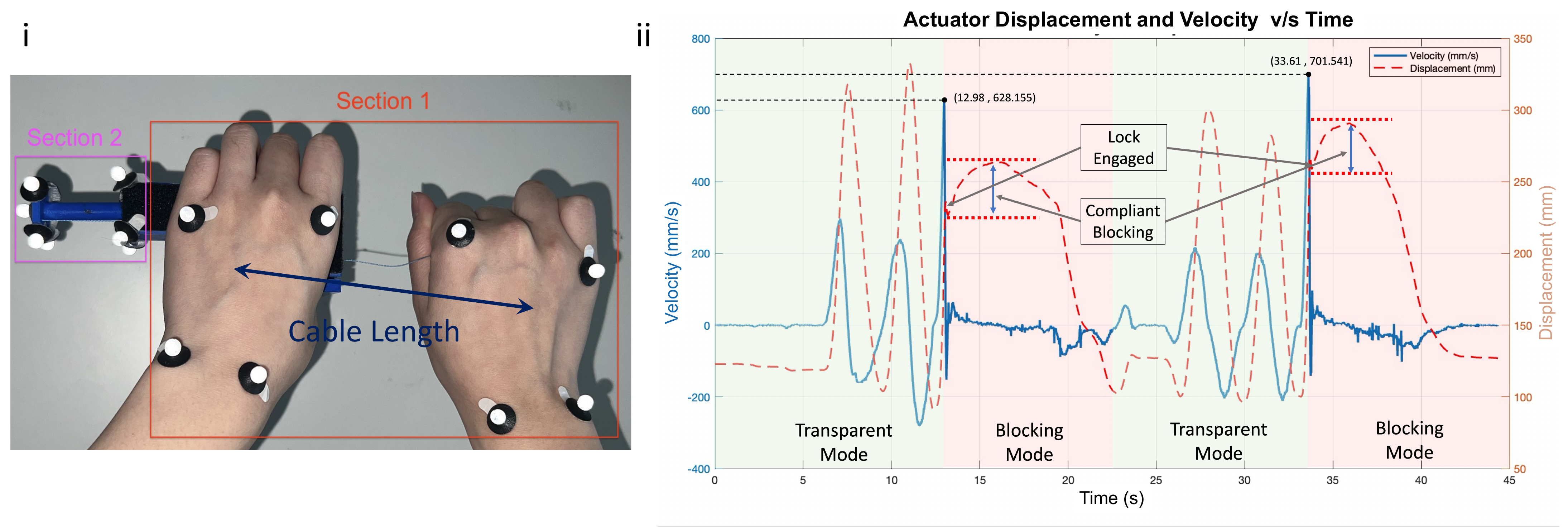}
    \caption{An overview of the (i). MoCap setup and (ii). results of preliminary validation of the prototype. The results show device transparency during low-to-medium cable velocities (green shaded regions), and the transition to blocking, and compliant blocking above a certain threshold (red shaded regions).}
    \label{fig:validationResults}
\end{figure*}

\section{Device Development and Validation}

\subsection{Working Principle and Prototyping}
The prototype (Fig.\,\ref{fig:summaryPic}) was conceptualised to be fixed to the wheelchair with a cable extending from it and anchoring to a fabric-based back-support-like vest worn by the user (Figs.\,\ref{fig:summaryPic}b,\ref{fig:deviceConcept}). An exploded view of the components of the developed prototype and bill of materials used is presented in Fig.\,\ref{fig:deviceConcept}. As introduced previously, the proposed torso stabiliser system needs to be non-restrictive during regular torso movements while engaging and locking as soon as a fall is detected. This resulted in the device requiring three distinct phases/modes:
    \subsubsection{Transparent Mode} In this mode, the cable connecting the device to the wheelchair maintains low tension to ensure the cable is taut at all times to track the user's movements while allowing for the user to feel unconstrained. This is achieved through a coil spring connected to the capstan, around which the cable is wound (Part\#2 and Part\#11 in Fig.\,\ref{fig:deviceConcept}).
    \subsubsection{Transparent-to-Blocking Mode Transition} The aim of this part of the device is to switch into the blocking mode when a fall event is detected, and reset afterwards. The developed mechanism (inset in Fig.\,\ref{fig:deviceConcept}) is based on the working principle used in speed governors, centrifugal clutches and seat-belts. In the developed mechanism, there are two pivoting flyweights (Part\#8) that move in and out as the capstan's angular velocity changes. The flyweights have Neodymium magnets (Part\#9), behaving analogously to springs or gravity in systems like centrifugal clutches or speed governors. The magnetic attraction force pulls the two flyweights inward, while increased angular velocity cause them to move outward. At the threshold angular velocity, the flyweights overcome the magnetic attraction completely and lock into the locking plate (Part\#6 and Part\#7), coupling the capstan to the blocking mode passive actuator and, by extension, creating a blocking force on the cable and the user. The choice of magnets and flyweight weights need to be identified iteratively based on experiments.   
    \subsubsection{Blocking Mode} This mode is triggered when the flyweights are engaged to the locking plate. The locking plate is connected via another cable to a nonlinear spring  (Part\#4 and Part\#5). Once the flyweights lock into the locking plate, the nonlinear spring couples with the main capstan, thereby, generating the blocking force. The nonlinear spring prevents abrupt blocking, and results in a compliant but definite stop.  

The off-the-shelf vest weighs $\approx300\,\si{\gram}$, and the developed device weighs $\approx130\,\si{\gram}$. The back-support vest was chosen as it has two strong plastic runner plates, that provide structural integrity to anchor the cable from the device. The cables were intended to be anchored as away from the torso as possible, to maximise tracking sensitivity. Most parts were designed and 3D-printed using FormLabs Form3+ printer with Tough 2000 resin, except for Locking Plate A (Part\#7) which was fabricated out of Aluminium using waterjet cutting. Even though they are load-bearing, the flyweights (Part\#8) continued to be in resin for this prototype, as the magnets inset within them provide additional structural integrity.

\subsection{MoCap-based Validation of Device Behaviour}

\subsubsection{Experimental Setup}
Preliminary validation for the prototyped device was conducted by pulling on the cable from the device with different velocities, while capturing data using the MoCap setup described previously. 
Fig.\,\ref{fig:validationResults}i shows the placement of markers on both the device and the two hands of the person performing this validation study. Four markers were placed on each body (the two hands, main body of the device and nonlinear spring plunger) to obtain each body's centroid and pose throughout the validation experiments, while ensuring redundancy.

\subsubsection{Results}
Results of the MoCap experiments are presented in Fig.\,\ref{fig:validationResults}ii. The blue and dashed orange lines represent cable velocity and displacement, respectively. At low-to-medium velocities (green shaded region), it can be clearly seen that the device cable promptly tracks extension and retraction, and is able to achieve non-restrictive travel of $55-60\,\si{\centi\meter}$. Additionally, as the velocity goes above a threshold of $\approx62.8\,\si[per-mode=symbol]{\centi\meter\per\second}$ (red shaded region), the blocking mechanism activates, and this results in a compliant but positive stop, as has been annotated in Fig.\,\ref{fig:validationResults}ii. The blocking mechanism resets as the tension in the cable is relieved. Two sets of blocking and reset operations are presented and annotated in the above trial. This MoCap study validates the concept's ability in having both non-restrictive motion as well as compliant blocking triggered above a certain cable velocity.

\section{Conclusion}
We addressed the challenge of torso stabilisation faced by individuals with SCI through the design, fabrication and validation of a purely mechanical device. As part of the problem statement, the device is required to be non-restrictive during ADL, as well as capable of compliant blocking in a prompt and responsive manner on detecting incipient fall. Through this work, we initially identified device design requirements such as unrestricted cable travel and blocking velocity. This was achieved through virtual prototyping of the potential device in OpenSim using MoCap data from experiments obtained from an individual with SCI performing leaning movements experienced during ADL. 

An early prototype of the proposed device was presented in the second part of this paper. A mechanical solution was designed to ensure robustness, simplicity, and cost-effectiveness, potentially facilitating adoption in low- and middle-income countries. Blocking is triggered by a mechanism inspired by centrifugal clutches and speed governors. Preliminary validation was conducted to understand the kinematic behaviour of the device and clearly demonstrate the prototype's ability to achieve non-restrictive travel, prompt transition to blocking mode, and ensure compliant but definite blocking. 

While we clearly identified device parameters, and early results with the developed prototype are promising, further work is required to tune the velocity at which the blocking mode is triggered. The developed prototype currently blocks at a lower velocity than was obtained from our simulation, causing false-positives. Further investigation is required to identify whether velocity should continue as the parameter of choice, or whether acceleration of the torso would be more apt. Finally, future prototypes need to be more robust, and validated with individuals without SCI, before a comprehensive trial can be conducted with individuals with SCI. 

\section*{Acknowledgement}
This work utilised expertise and prototyping equipment at the Imperial College Advanced Hackspace. The authors would like to thank the mentors at the hackspace for their support.

\bibliographystyle{IEEEtran}
\bibliography{IEEEabrv,references}

\end{document}